\documentclass[twoside]{article}

\usepackage[accepted]{aistats2014}
\usepackage[colorinlistoftodos]{todonotes}
\usepackage{tikz}
\usepackage{pgfplots}
\usetikzlibrary{arrows,backgrounds,fit,positioning}
\tikzstyle{int}=[draw, minimum size=1em]
\tikzstyle{init} = [pin edge={to-,thin,black}]
\tikzstyle{surround} = [draw, minimum size=2em]
\usepackage[framemethod=TikZ]{mdframed}
\usepackage{comment}
\usepackage{caption}
\usepackage{subcaption}
\usepackage{amsmath}
\usepackage{mynotation}
\usepackage{dsfont}
\usepackage{wrapfig}
\usepackage{lipsum}
\usepackage{algorithm}

\newcommand{\myargmin}[2]{\underset{#1}{\mathrm{arg}\,\mathrm{min}}\,\,#2}
\newcommand{\feature}{z}
\newcommand{\featurebig}{Z}
\newcommand{\meas}{y}

\begin{document}

\twocolumn[

\aistatstitle{Learning deep dynamical models from image pixels}

\aistatsauthor{Niklas Wahlstr\"om$^1$ \And Thomas B. Sch\"on$^2$ \And Marc Peter Deisenroth$^3$}

\aistatsaddress{\\ $^1$Department of Electrical Engineering, Link\"oping
  University, Sweden \\
 $^2$Department of Information Technology, Uppsala
  University, Sweden \\
$^3$Department of Computing, Imperial College
  London, United Kingdom } 

]

\begin{abstract}
Modeling dynamical systems is important in many disciplines, e.g., control, robotics, or neurotechnology. Commonly the state of these systems is not directly observed, but only available through noisy and potentially high-dimensional observations. In these cases, system identification, i.e., finding the measurement mapping and the transition mapping (system dynamics) in latent space can be challenging. For linear system dynamics and measurement mappings efficient solutions for system identification are available. However, in practical applications, the linearity assumptions does not hold, requiring non-linear system identification techniques. If additionally the observations are high-dimensional (e.g., images), non-linear system identification is inherently hard. To address the problem of non-linear system identification from high-dimensional observations, we combine recent advances in deep learning and system identification. In particular, we jointly learn a low-dimensional embedding of the observation by means of deep auto-encoders and a predictive transition model in this low-dimensional space. We demonstrate that our model enables learning good predictive models of  dynamical systems from pixel information only.
\end{abstract}

\section{Introduction}
High-dimensional time series include video streams, Electroencephalography (EEG) and sensor network data. Dynamical models describing such data are desired for forecasting (prediction) and controller design, both of which play an important role, e.g., in autonomous systems, machine translation, robotics and surveillance applications. A key challenge is system identification, i.e., finding a mathematical model of the dynamical system based on the information provided by measurements from the underlying system. In the context of  state-space models this includes finding two functional relationships between (a) the states at different time steps (prediction\slash transition model) and (b) states and corresponding measurements (observation\slash measurement model). In the linear case, this problem is very well studied, and many standard techniques exist, e.g., subspace methods~\cite{deMoor:1996}, expectation maximization~\cite{ShumwayS:1982,Ghahramani:1998} and  prediction-error methods~\cite{Ljung:1999}. However, in realistic and practical scenarios we require non-linear system identification techniques. 

Learning non-linear dynamical models is an inherently difficult problem, and it has been one of the most active areas in system identification for the last decades~\cite{Ljung2010,Sjoberg:1995}. In recent years, sequential Monte Carlo (SMC) methods have received attention for identifying non-linear state-space models~\cite{Schon2011}, see also the recent surveys~\cite{KantasDSM:2009}. While methods based on SMC are powerful, they are also computationally expensive.
Learning non-linear dynamical models from very high-dimensional sensor data is even more  challenging. First, finding (non-linear) functional relationships in very high dimensions is hard (un-identifiability, local optima, overfitting, etc.); second, the amount of data required to find a good function approximator is enormous.
Fortunately, high-dimensional data often possesses an intrinsic lower dimensionality. We will exploit this property for system identification by finding a low-dimensional representation of high-dimensional data and learning predictive models in this low-dimensional space. For this purpose, we need an automated procedure for finding compact low-dimensional representations\slash features.

The state of the art in learning parsimonious representations of high-dimensional data is currently defined by deep learning architectures, such as deep neural networks~\cite{Hinton2006}, stacked\slash deep auto-encoders~\cite{Vincent2008} and convolutional neural networks~\cite{LeCun1998}, all of which have been successfully applied to image, text, speech and audio data in commercial products, e.g., by Google, Amazon and Facebook. Typically, these feature learning methods are applied to static data sets, e.g., for image classification. The auto-encoder gives explicit expressions of two generative mappings: 1) an encoder $g\inv$ mapping the high-dimensional data to the features, and 2) a decoder $g$ mapping the features to high-dimensional reconstructions.
In the machine learning literature, there exists a vast number of other well studied nonlinear dimensionality reduction methods such as the Gaussian process latent variable model (GP-LVM) \cite{Lawrence2005}, kernel PCA~\cite{Schoelkopf1998}, Laplacian Eigenmaps~\cite{Belkin2003} and Locally Linear Embedding \cite{Roweis:2000}. However, all of them provide at most one of the two mappings $g$ and $g\inv$.

\begin{figure*}[tb]
\centering
\includegraphics[width = \hsize]{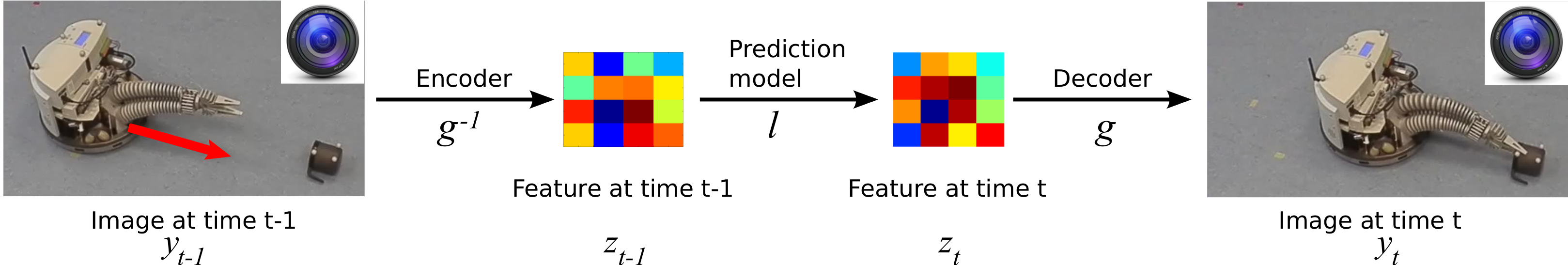}
\caption{Combination of deep learning architectures for feature learning and predictor models in feature space. A camera observes a robot approaching an object. A good low-dimensional feature representation of an image is important for learning a predictive model if the camera is the only sensor available.}
\label{fig:model illustration}
\end{figure*}
%
In this paper, we combine feature learning and dynamical systems modeling to obtain good predictive models for high-dimensional time series. In particular, we use deep auto-encoder neural networks for automatically finding a compact low-dimensional representation of an image. In this low-dimensional feature space, we use a neural network for modeling the nonlinear system dynamics. The embedding and the predictive model in feature space are learned \emph{jointly}. An simplified illustration of our approach is shown in \fig\ref{fig:model illustration}. An encoder $g\inv$ maps an image $\meas_{t-1}$ at time step $t-1$ to a low-dimensional feature $\feature_{t-1}$. In this feature space, a prediction model $l$ maps the feature forward in time to $\feature_t$. Subsequently, the decoder $g$ can be used to generate a predicted image $\meas_t$ at the next time step. This framework needs access to both the encoder $g\inv$ and the decoder $g$, which motivates our use of the auto-encoder as dimensionality reduction technique. 

Consequently, the contributions of this paper are (a) a model for learning a low-dimensional dynamical representation of high-dimensional data, which can be used for long-term predictions; (b) experimental evidence demonstrating that jointly learning the parameters of the latent embedding and  the predictive model in latent space can increase the performance compared to a separate training.

\section{Model}
\label{sec:model}
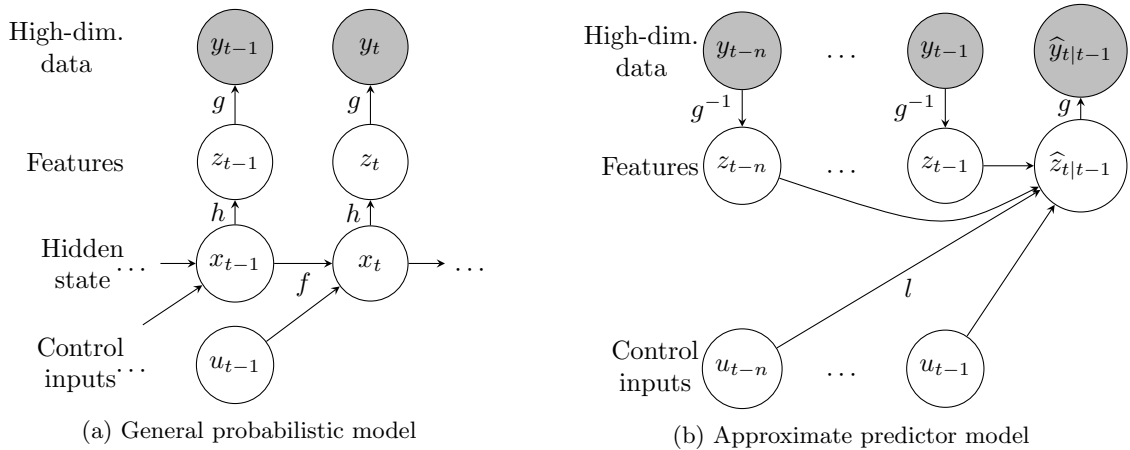
\begin{figure*}[t]
\begin{subfigure}{0.46\linewidth}
\tikzset{%
  every neuron/.style = {
    circle,
    draw,
    minimum size=1cm
  },  
	neuron missing/.style={
		draw=none, 
		scale=1,
		text height=0.333cm,
		execute at begin node=\color{black}$\cdots$
	},
}

\centering
\begin{tikzpicture}[x=0.9cm, y=0.9cm, >=stealth]
\node [every neuron/.try] (x1) at (-1,0) {$x_{t-1}$};
\node [every neuron/.try] (x2) at (1,0) {$x_{t}$};
\node [neuron missing/.try] (x0) at (-2.5,0) {};
\node [neuron missing/.try] (d) at (-2.5,-1.5) {};
\node [neuron missing/.try] (x3) at (2.5,0) {};
\node [every neuron/.try] (y1) at (-1,1.5) {$\feature_{t-1}$};
\node [every neuron/.try] (y2) at (1,1.5) {$\feature_{t}$};
\node [every neuron/.try,fill=lightgray] (z1) at (-1,3.2) {$\meas_{t-1}$};
\node [every neuron/.try,fill=lightgray] (z2) at (1,3.2) {$\meas_{t}$};
\node [] (u0) at (-2.5,-1) {};
\node [every neuron/.try] (u1) at (-1,-1.5) {$u_{t-1}$};
\node [align=center, left] at (-2.5,3.2) {High-dim.\\ data}; 
\node [align=center, left] at (-2.5,1.5) {Features}; 
\node [align=center, left] at (-2.5,0) {Hidden \\state}; 
\node [align=center, left] at (-2.5,-1.5) {Control \\ inputs}; 
\draw [->] (x1) -- (x2) node[midway,below] {$f$};
\draw [->] (x0) -- (x1);
\draw [->] (x2) -- (x3);
\draw [->] (x1) -- (y1) node[midway,left] {$h$};
\draw [->] (y1) -- (z1) node[midway,left] {$g$};
\draw [->] (x2) -- (y2) node[midway,left] {$h$};
\draw [->] (y2) -- (z2) node[midway,left] {$g$};
\draw [->] (u0) -- (x1);
\draw [->] (u1) -- (x2);
\end{tikzpicture}
\caption{General probabilistic model}
\label{fig:probabilistic_model}
\end{subfigure}
\begin{subfigure}{0.46\linewidth}
\tikzset{%
  every neuron/.style = {
    circle,
    draw,
    minimum size=1cm
  },  
	neuron missing/.style={
		draw=none, 
		scale=1,
		text height=0.333cm,
		execute at begin node=\color{black}$\cdots$
	},
}
\centering
\begin{tikzpicture}[x=0.9cm, y=0.9cm, >=stealth]
\node [every neuron/.try] (u0) at (-4,-1.5) {$u_{t-n}$};
\node [every neuron/.try] (u1) at (-1,-1.5) {$u_{t-1}$};
\node [every neuron/.try] (y0) at (-4,1.5) {$\feature_{t-n}$};
\node [every neuron/.try] (y1) at (-1,1.5) {$\feature_{t-1}$};
\node [every neuron/.try] (y2) at (1,1.5) {$\widehat{\feature}_{t\mid t-1}$};
\node [neuron missing/.try] (d1) at (-2.5,-1.5) {};
\node [neuron missing/.try] (d2) at (-2.5,1.5) {};
\node [neuron missing/.try] (d3) at (-2.5,3.2) {};
\node [every neuron/.try,fill=lightgray] (z0) at (-4,3.2) {$\meas_{t-n}$};
\node [every neuron/.try,fill=lightgray] (z1) at (-1,3.2) {$\meas_{t-1}$};
\node [every neuron/.try,fill=lightgray] (z2) at (1,3.2) {$\widehat{\meas}_{t\mid t-1}$};
\node [align=center, left] at (-4.5,3.2) {High-dim. \\ data};
\node [align=center, left] at (-4.5,1.5) {Features}; 
\node [align=center, left] at (-4.5,-1.5) {Control \\ inputs}; 
\draw [->] (y1) -- (y2);
\draw [->] (u1) -- (y2);
\draw [->] (u0) -- (y2) node[midway,below] {$l$};
\draw [->] (z1) -- (y1) node[midway,left] {$g^{-1}$};
\draw [->] (z0) -- (y0) node[midway,left] {$g^{-1}$};
\draw [->] (y2) -- (z2) node[midway,left] {$g$};
\draw[->] (y0) .. controls (-1,0.5) .. (y2);
\end{tikzpicture}
\caption{Approximate predictor model}
\label{fig:predictor_model}
\end{subfigure}
\caption{(\subref{fig:probabilistic_model}) The general graphical model of the high-dimensional data $\meas_t$. Each data point $\meas_t$ has a low-dimensional representation $\feature_t$, which is modeled using a state-space model with hidden state $x_t$ and control input $u_t$.  (\subref{fig:predictor_model}) The approximate predictor model, where the predicted feature $\widehat{\feature}_{t\mid t-1}$ is a function of the $n$ past features $\feature_{t-n}$ to $\feature_{t-1}$ and $n$ past control inputs $u_{t-n}$ to $u_{t-1}$. Each of the features $\feature_{t-n}$ to $\feature_{t-1}$ is computed from high-dimensional data $\meas_{t-n}$ to $\meas_{t-1}$ via the encoder $g^{-1}$. The predicted feature $\widehat{\feature}_{t\mid t-1}$ is mapped to predicted high-dimensional data via the decoder $g$.}
\end{figure*}

We consider a dynamical system where control inputs are denoted by $u$ and observations are denoted by $\meas$. In the context of this paper, the observations are pixel information from images. We assume that a low-dimensional latent variable $\feature$ exists that compactly represents the relevant properties of $\meas$.
Since we consider dynamical systems, a low-dimensional representation $\feature$ of a (static) image $\meas$ is insufficient to capture important dynamic information, such as velocities. Therefore, we introduce an additional latent variable $x$, the \emph{state}. In our case, the state $x_t$ contains features from multiple time steps (e.g., $t-1$ and $t$) to capture velocity (or higher-order) information. Therefore, our transition model does not map features at time $t-1$ to time $t$ (as illustrated in \fig\ref{fig:model illustration}), but the transition function $f$ maps states $x_{t-1}$ (and controls $u_{t-1}$) to states $x_t$ at time $t$. The full dynamical system is given as the state-space model
\begin{subequations}
 \label{eq:model}
\begin{align}
x_{t+1} & = f(x_t,u_t;\theta) + w_t(\theta), \label{eq:model_x}\\
\feature_t     & = h(x_t;\theta) + v_t(\theta), \label{eq:model_y}\\
\meas_t     & = g(\feature_t;\theta) + e_t(\theta), \label{eq:model_z}
\end{align}
\end{subequations}
where each measurement $\meas_t$ can be described by a low-dimensional feature representation $\feature_t$ \eqref{eq:model_z}. These features are in turn modeled with a low-dimensional state-space model in \eqref{eq:model_x} and \eqref{eq:model_y}, where the state $x_t$ contains the full information about the state of the system at time instant~$t$, see also \fig\ref{fig:probabilistic_model}. Here $w_t(\theta)$, $v_t(\theta)$ and $e_t(\theta)$ are sequences of independent random variables and $\theta$ are the model parameters.

\subsection{Approximate predictor model}
\label{sec:predictor_model}
To identify parameters in dynamical systems, the prediction-error method \cite{Ljung:1999} has been applied extensively within the system identification community during the last five decades. It is based on minimizing the error between the sequence of measurements $\meas_t$ and the predictions $\widehat{\meas}_{t\mid t-1}(\theta)$, usually the one-step ahead prediction. To achieve this, we need a predictor model that relates the prediction $\widehat{\meas}_{t\mid t-1}(\theta)$ to all previous measurements, control inputs and the system parameters $\theta$.

In general, it is difficult to derive a predictor model based on the nonlinear state-space model \eqref{eq:model}, and a closed form expression for the prediction is only available in a few special cases~\cite{Ljung:1999}. However, by approximating the optimal solution, a predictor model for the features $\feature_t$ can be stated in the form
\begin{align}
\label{eq:PM}	
\widehat{\feature}_{t\mid t-1}(\theta_{\text{M}}) = l(\featurebig_{t-1};\theta_{\text{M}}),
\end{align}
where $\featurebig_{t-1} = (\feature_1,u_1,\dots,\feature_{t-1},u_{t-1})$ includes all past features and control inputs,  $l$ is a nonlinear function and $\theta_{\text{M}}$ is the corresponding model parameters. Note that the predictor model no longer has an explicit notion of the state $x_t$. The model introduced in~\eqref{eq:PM} is indeed very flexible, and in this work we have restricted this flexibility somewhat by working with a nonlinear autoregressive exogenous model (NARX) \cite{Ljung:1999}, which relates the predicted current value $\widehat{\feature}_{t\mid t-1}(\theta_{\text{M}})$ of the time series to the past $n$ values of the time series $\feature_{t-1}, \feature_{t-2},\dots, \feature_{t-n}$, as well as the past $n$ values of the control inputs $u_{t-1}, u_{t-2},\dots, u_{t-n}$. The resulting NARX predictor model is given by
\begin{align}
\widehat{\feature}_{t\mid t-1}(\theta_{\text{M}}) = l(\feature_{t-1},u_{t-1},\dots,\feature_{t-n},u_{t-n};\theta_{\text{M}}),
\end{align}
where 
$l$ is a nonlinear function, in our case a neural network. The model parameters in the nonlinear function are normally estimated by minimizing the sum of the prediction errors $\feature_t - \widehat{\feature}_{t\mid t-1}(\theta_{\text{M}})$.
However, since we are interested in a good predictive performance for the high-dimensional data $\meas$ rather than for the features $\feature$, we transform the predictions back to the high-dimensional space and obtain a prediction $\widehat{\meas}_{t|t-1} = g(\widehat{\feature}_{t|t-1};\theta_{\text{D}})$, which we use in our error measure.

An additional complication is that we do not have access to the features $\feature_t$. 
Therefore, before training, the past values of the time series have to be replaced with their feature representation $y = g^{-1}(\meas;\theta_{\text{E}})$, which we compute from the pixel information $\meas$. Here, $g^{-1}$ is an approximate inverse of $g$, which will be described in more detail the next section. This gives the final predictor model
\begin{subequations}
\begin{align} \label{eq:prediction}
&\widehat{\meas}_{t\mid t-1}(\theta_{\text{E}},\theta_{\text{D}},\theta_{\text{M}})  = g(\widehat{\feature}_{t\mid t-1}(\theta_{\text{E}},\theta_{\text{M}});\theta_{\text{D}}), \\
&\widehat{\feature}_{t\mid t-1}(\theta_{\text{E}},\theta_{\text{M}})				   = l(\feature_{t-1}(\theta_{\text{E}}),u_{t-1},\dots,\feature_{t-n}(\theta_{\text{E}}),u_{t-n};\theta_{\text{M}}), \\
&\feature_t(\theta_{\text{E}}) 											   = g^{-1}(\meas_{t};\theta_{\text{E}}), \label{eq:prediction_mapping}
\end{align}
\end{subequations}
which is also illustrated in \fig\ref{fig:predictor_model}. The corresponding prediction error will be
\begin{align} \label{eq:prediction_error}
\varepsilon_t^{\text{P}}(\theta_{\text{E}},\theta_{\text{D}},\theta_{\text{M}}) = \meas_t - \widehat{\meas}_{t\mid t-1}(\theta_{\text{E}},\theta_{\text{D}},\theta_{\text{M}}).
\end{align}

\subsection{Auto-encoder}
\label{sec:autoencoder}

\begin{figure}
\centering
\tikzset{%
  every neuron/.style={
    circle,
    draw,
    minimum size=0.7cm    
  },
  neuron missing/.style={
    draw=none, 
    fill=none, 
    scale=1,
    text height=0.333cm,
    execute at begin node=\color{black}$\vdots$
  },
}

\begin{tikzpicture}[x=0.7cm, y=1cm, >=stealth]

\foreach \m/\l [count=\y] in {1,2,3,missing,4}
  \node [every neuron/.try, fill=lightgray, neuron \m/.try] (input-\m) at (0,2.5-\y) {};

\foreach \m [count=\y] in {1,missing,2}
  \node [every neuron/.try, neuron \m/.try ] (hiddena-\m) at (2,2-\y*1.25) {};

\foreach \m [count=\y] in {1,missing,2}
  \node [every neuron/.try, neuron \m/.try ] (hiddenb-\m) at (4,1.5-\y) {};
	
\foreach \m [count=\y] in {1,missing,2}
  \node [every neuron/.try, neuron \m/.try ] (hiddenc-\m) at (6,2-\y*1.25) {};
	
\foreach \m/\l [count=\y] in {1,2,3,missing,4}
  \node [every neuron/.try, fill=lightgray, neuron \m/.try] (output-\m) at (8,2.5-\y) {};

\foreach \l [count=\i] in {1,2,3,M}
  \draw [<-] (input-\i) -- ++(-1.2,0)
    node [above, midway] {$\meas_{\l,t}$};

\foreach \l [count=\i] in {1,m}
  \node [above] at (hiddenb-\i.north) {$\feature_{\l,t}$};

\foreach \l [count=\i] in {1,2,3,M}
  \draw [->] (output-\i) -- ++(1.9,0)
    node [above, midway] {$\widehat{\meas}_{\l,t\mid t}$};

\foreach \i in {1,...,4}
  \foreach \j in {1,...,2}
    \draw [->] (input-\i) -- (hiddena-\j);

\foreach \i in {1,...,2}
  \foreach \j in {1,...,2}
    \draw [->] (hiddena-\i) -- (hiddenb-\j);
\foreach \i in {1,...,2}
  \foreach \j in {1,...,2}
    \draw [->] (hiddenb-\i) -- (hiddenc-\j);
		
\foreach \i in {1,...,2}
  \foreach \j in {1,...,4}
    \draw [->] (hiddenc-\i) -- (output-\j);

\foreach \l [count=\x from 0] in {Input layer \\(high-dim. data), , Hidden layer \\ (feature), , Output layer \\ (reconstructed)}
  \node [align=center, below] at (\x*2,3) {\l}; 
	\foreach \l [count=\x from 0] in { ,$\underbrace{\qquad\qquad\qquad}_{\text{Encoder }g^{-1}}$ , , $\underbrace{\qquad\qquad\qquad}_{\text{Decoder }g}$ , }
  \node [align=center, below] at (\x*2,-2.8) {\l}; 
\end{tikzpicture}
\caption{An auto-encoder consisting of an encoder $g\inv$ and a decoder $g$. The original image $\meas_t = [\meas_{1,t},\cdots,\meas_{M,t}]^\Transp$ is  mapped to its low-dimensional representation $\feature_t = [\feature_{1,t},\cdots,\feature_{m,t}]^\Transp = g\inv(\meas_t)$ with the encoder, and then back to a high-dimensional representation $\widehat{\meas}_{t\mid t-1} = g(\widehat{\feature}_{t\mid t-1})$ by the decoder $g$, where $M \gg m$.}
\label{fig:autoencoder}
\end{figure}
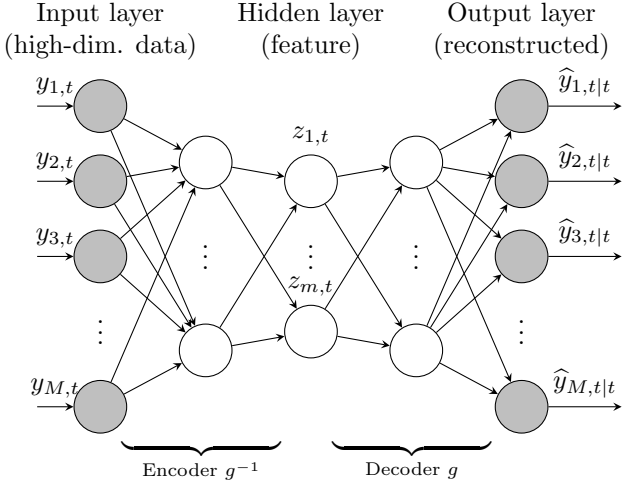
%
We use a deep auto-encoder neural network to parameterize the feature mapping and its inverse. It consists of a deep encoder network $g^{-1}$ and a deep decoder network $g$.
Each layer $k$ of the encoder neural network $g^{-1}$ computes $\meas^{(k+1)}_t = \sigma(A_k \meas^{(k)}_t + b_k)$, where $\sigma$ is a squashing function and $A_k$ and $b_k$ are free parameters. The control input to the first layer is the image, i.e., $\meas^{(1)}_t = \meas_t$. The last layer is the low-dimensional feature representation of the image $\feature_t(\theta_{\text{E}}) = g^{-1}(\meas_t;\theta_{\text{E}})$, where $\theta_{\text{E}} = [\dots, A_k, b_k,\dots]$ are the parameters of all neural network layers.
The decoder $g$ consists of the same number of layers in reverse order, see \fig\ref{fig:autoencoder}. It can be considered an approximate inverse of the encoder, such that $\widehat{\meas}_{t\mid t}(\theta_{\text{E}},\theta_{\text{D}}) \approx \meas_t$, where 
\begin{align}
\widehat{\meas}_{t\mid t}(\theta_{\text{E}},\theta_{\text{D}}) = g(g^{-1}(\meas_t;\theta_{\text{E}});\theta_{\text{D}})
\end{align}
is the reconstructed version of $\meas_t$.
In the classical setting, the encoder and the decoder are trained simultaneously to minimize the reconstruction error 
\begin{align} \label{eq:reconstruction_error}
\varepsilon_t^{\text{R}}(\theta_{\text{E}},\theta_{\text{D}}) = \meas_t - \widehat{\meas}_{t\mid t}(\theta_{\text{E}},\theta_{\text{D}}),
\end{align}
where the parameters of $g$ and $g^{-1}$ optionally can be coupled to constrain the solution to some degree~\cite{Vincent2008}. 

We realize that the autoencoder suits our problem at hand very well, since it provides an explicit expression of both the mapping from features to data $g$ as well as its approximate inverse $g^{-1}$, which is convenient to form the predictions in~\eqref{eq:prediction}. Many other nonlinear dimensionality reduction methods such as the Gaussian process latent variable model (GP-LVM) \cite{Lawrence2005}, kernel PCA~\cite{Schoelkopf1998}, Laplacian Eigenmaps~\cite{Belkin2003} and Locally Linear Embedding \cite{Roweis:2000} do not provide an explicit expression of both mappings $g$ and $g\inv$. This is the main motivation way we use the (deep) auto-encoder model for dimensionality reduction.
%

\section{Training}
\label{sec:training}
To summarize, our model contains the following free parameters: the parameters for the encoder $\theta_{\text{E}}$, the parameters for the decoder $\theta_{\text{D}}$ and the parameters for the predictor model $\theta_{\text{M}}$. To train the model, we employ two cost functions, the sum of the prediction errors~\eqref{eq:prediction_error}, 
\begin{subequations}
\begin{align}
V_{\text{P}}(\theta_{\text{E}}, \theta_{\text{D}},\theta_{\text{M}}) 
 & = \log\left(\frac{1}{2NM}\sum_{t=1}^N \| \varepsilon_t^{\text{P}}(\theta_{\text{E}},\theta_{\text{D}},\theta_{\text{M}})\|^2\right),
 \label{eq:training_objective1}
\end{align}
and the sum of the reconstruction errors~\eqref{eq:reconstruction_error},
\begin{align}
V_{\text{R}}(\theta_{\text{E}},\theta_{\text{D}}) 
& = \log\left(\frac{1}{2NM}\sum_{t=1}^N \|\varepsilon_t^{\text{R}}
(\theta_{\text{E}},\theta_{\text{D}})\|^2\right).
\label{eq:training_objective2}
\end{align}
\end{subequations}

Generally, there are two ways of finding the model parameters:
(1) separate training and (2) joint training, both of which are explained below.

\subsection{Separate training}
\label{sec:separate_training}
Normally when features are used for inference of dynamical models, they are first extracted from the data in a pre-processing step and as a second step the predictor model is estimated based on these features.  In our setting, this would correspond to sequentially training the model with using two cost functions \eqref{eq:training_objective1}--\eqref{eq:training_objective2} where we first learn a compact feature representation by minimizing the reconstruction error
\begin{subequations}
\label{eq:separate_training}
\begin{align}
\big(\widehat{\theta}_{\text{E}},\widehat{\theta}_{\text{D}}\big) & \in \myargmin{\theta_{\text{E}},\theta_{\text{D}}}{V_{\text{R}}(\theta_{\text{E}},\theta_{\text{D}})},
\label{eq:separate_training1}
\end{align}
and subsequently minimize the prediction error 
\begin{align}
\widehat{\theta}_{\text{M}} & = \myargmin{\theta_{\text{M}}}{V_{\text{P}}(\widehat{\theta}_{\text{E}}, \widehat{\theta}_{\text{D}},\theta_{\text{M}})}
\label{eq:separate_training2}
\end{align}
\end{subequations}
with fixed auto-encoder parameters $\widehat{\theta}_{\text{E}}, \widehat{\theta}_{\text{D}}$.
The gradients of these cost functions with respect to the model parameters can be computed efficiently by back-propagation. The cost functions are then minimized by the BFGS algorithm~\cite{NocedalW:2006}.

\subsection{Joint training}
An alternative to separate training is to minimize the reconstruction error and the prediction error simultaneously by considering the optimization problem
\begin{align} 
\label{eq:joint_training}
\big(\widehat{\theta}_{\text{E}},\widehat{\theta}_{\text{D}},\widehat{\theta}_{\text{M}}\big)= 
\myargmin{\theta_{\text{E}},\theta_{\text{D}},\theta_{\text{M}}}{\left(V_{\text{R}}(\theta_{\text{E}},\theta_{\text{D}}) 
+  V_{\text{P}}(\theta_{\text{E}},\theta_{\text{D}},\theta_{\text{M}}) \right)},
\end{align}
where we jointly optimize the free parameters in both the auto-encoder $\theta_{\text{E}}$, $\theta_{\text{D}}$ and the predictor model $\theta_{\text{M}}$. Again, back-propagation is used for computing the gradients of this cost function. 

\subsection{Initialization}
\label{sec:pretraining}
The auto-encoder has strong similarities with principal component analysis (PCA). More precisely, if we use a linear activation function and only consider a single layer, the auto-encoder and PCA are identical \cite{Bourlard1988}. We exploited this relationship to initialize the parameters of the auto-encoder. The auto-encoder network has been unfolded, each pair of layers in the encoder and the decoder have been combined, and the corresponding PCA solution has been computed for each of these pairs. By starting with the high-dimensional data at the top layer and using the principal components from that pair of layers as input to the next pair of layers, we recursively compute a good initialization for all parameters in the auto-encoder network. Similar pre-training routines are found in \cite{Hinton2006} where a restricted Boltzmann machine is used instead of PCA.

\section{Results}
\label{sec:results}

We report results on identification of (1) the nonlinear dynamics of a pendulum (1-link robot arm) moving in a horizontal plane and the torque as control input, (2) an object moving in the 2D-plane, where the 2D velocity serves as control input. In both examples, we learn the dynamics solely based on pixel information. Each pixel $\meas_t^{(i)}$ is a component of the measurement $\meas_t = [\meas_t^{(1)}, \dots, \meas_t^{(M)}]^\Transp$ and assumes a continuous gray-value in $[0,1]$.

\begin{figure*}[t]
\centering
\begin{subfigure}{0.46\linewidth}
	\includegraphics[width = \linewidth]{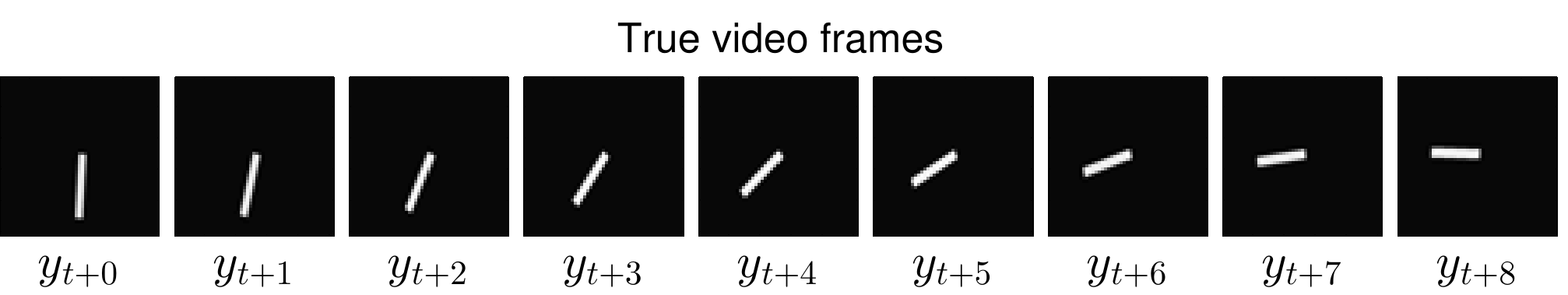}
    \includegraphics[width = \linewidth]{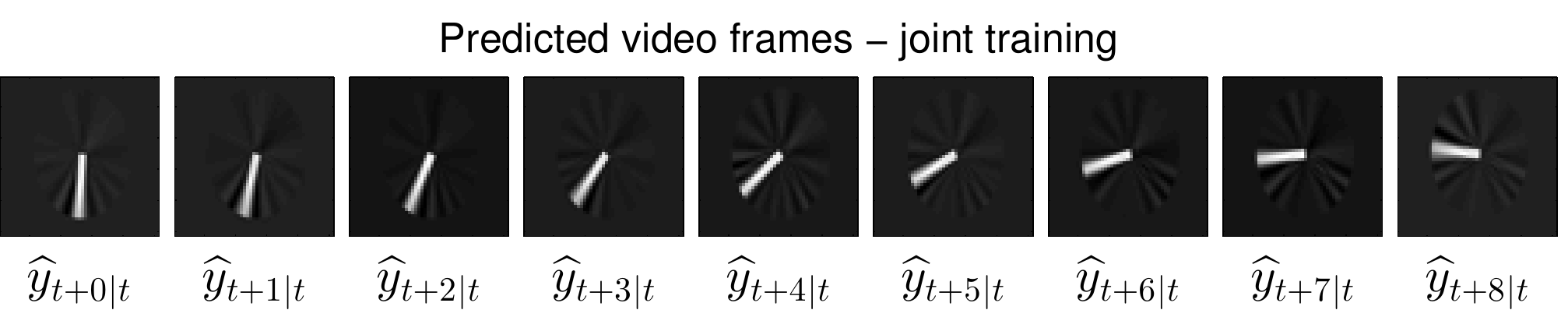}
    \includegraphics[width = \linewidth]{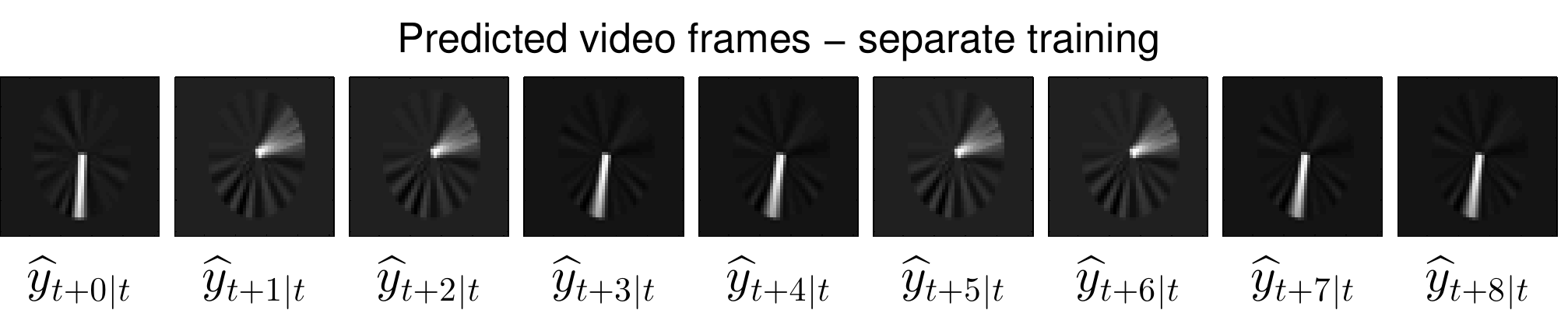}
	\caption{Sequence 1}
    \label{fig:results1}
\end{subfigure}
\hfill
\begin{subfigure}{0.46\linewidth}
    \includegraphics[width = \linewidth]{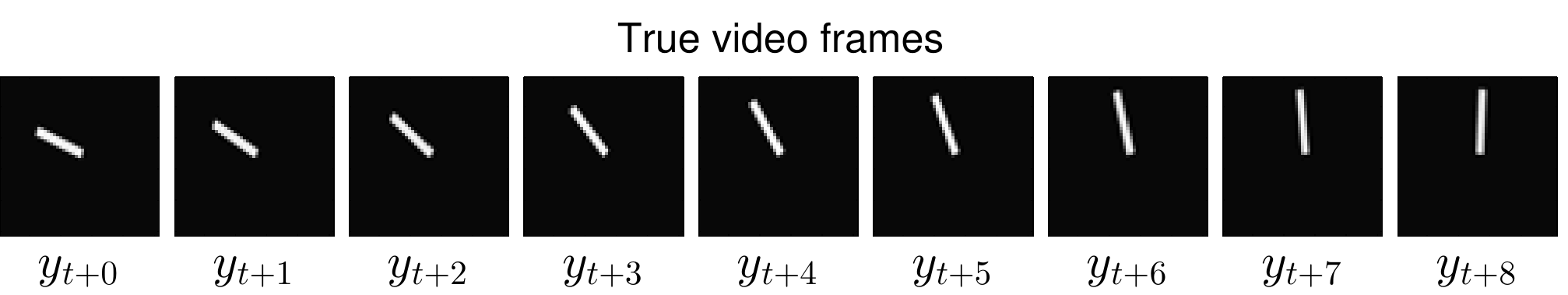}
    \includegraphics[width = \linewidth]{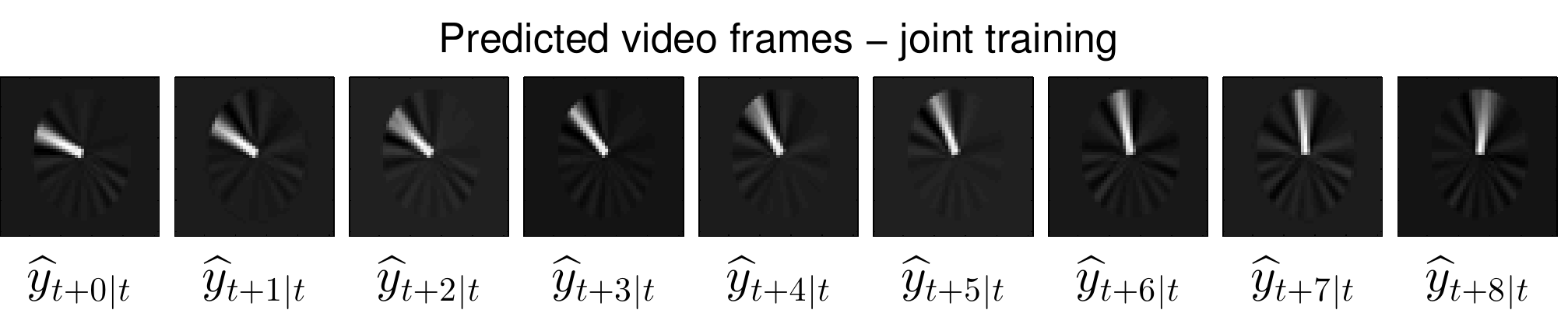}
    \includegraphics[width = \linewidth]{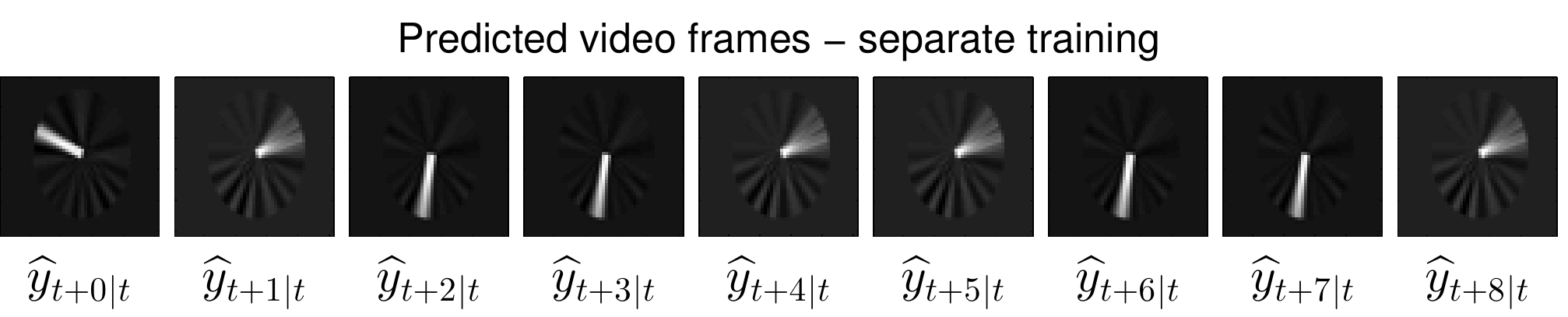}
	\caption{Sequence 2}
	\label{fig:results2}
\end{subfigure}
\caption{
Exp. 1: Two typical image sequences and corresponding prediction results (validation data), computed according to \eqref{eq:multi_step_prediction}. 
The top rows show nine consecutive ground truth image frames from time instant $t$ to $t+8$. The second and the third rows display the corresponding long-term ahead predictions based on measured images up to time $t$ for both joint  (center) and separate training (bottom) of the model parameters.} 
\label{fig:results}
\end{figure*}

\subsection{Experiment 1: Pendulum in the plane}
We simulated 400 frames of a pendulum moving in a plane with $51 \times 51 =2\thinspace601$ pixels in each frame and the torque of the pendulum as control input. To speed up training, the image input has been reduced to $\text{dim}(\meas_t) = 50$ prior to model learning (system identification) using PCA. With these 50 dimensional inputs, four layers have been used for the encoder $g^{-1}$ as well as the decoder $g$ with dimension 50-25-12-6-2. Hence, the features have dimension $\text{dim}(x_t) = 2$. The order of the dynamics was chosen as $n = 2$ to capture velocity information. For the predictor model $l$ we used a two-layer neural network with a 6-4-2 architecture.

We evaluate the performance in terms of long term predictions. These predictions are constructed by concatenating multiple 1-step ahead predictions. More precisely,  the $p$-step ahead prediction $\widehat{\meas}_{t+p|t} = g(\widehat{\feature}_{t+p|t})$ is computed iteratively as
\begin{subequations} \label{eq:multi_step_prediction}
\begin{align}
\widehat{\feature}_{t+1|t} & = l(\widehat{\feature}_{t|t},u_{t},\dots),\\
& \hdots \notag \\
\widehat{\feature}_{t+p|t} & = l(\widehat{\feature}_{t+p-1|t},u_{t+p-1|t},\dots),
\end{align}
\end{subequations}
where $\widehat{\feature}_{t|t} = g^{-1}(\meas_t)$ is the features of the data point at time instance $t$. We assumed that the applied future torques were known.

The predictive performance on two exemplary image sequences of the validation data  of our system identification models  is illustrated in \fig\ref{fig:results}, where control inputs (torques) are assumed known. The top rows show the ground truth images, the center rows show the predictions based on a model using joint training~\eqref{eq:joint_training}, the bottom rows show the corresponding predictions of a model where the auto-encoder and the predictive model were trained sequentially according to~\eqref{eq:separate_training}.
For the model based on jointly training all parameters, we obtain good predictive performance for both one-step ahead prediction and multiple-step ahead prediction. In contrast, the predictive performance of learning the features and the dynamics separately is worse than the predictive performance of the model trained by jointly optimizing all parameters. 
Although the auto-encoder does a perfect job (left-most frame, 0-step ahead prediction), already the (reconstructed) one-step ahead prediction is not similar to the ground-truth image. 
This can also be seen in Table~\ref{tab:result} where the reconstruction error is equally good for both models, but for the prediction error we manage to get a better value using joint training than using separate training.
Let us have a closer look at the model based on separate training: Since the auto-encoder performs well, the learned transition model is the reason for bad predictive performance. We believe that the auto-encoder found a good feature representation for reconstruction, but this representation was not ideal for learning a transition model.

\begin{table}[tb]
  \begin{center}
	\caption{Exp. 1: Prediction error $V_{\text{P}}$ and reconstruction error $V_{\text{R}}$ for separate and joint training.}
		\label{tab:result}
    \begin{tabular}{ l | l l}
			 Training & $V_{\text{P}}$ & $V_{\text{R}}$ \\
    \hline
			Joint training \eqref{eq:joint_training} & -6.91 & -6.92 \\
			Separate training \eqref{eq:separate_training} & -5.12 & -6.99 \\
    \end{tabular}
  \end{center}  
  \vspace{-5mm}
\end{table}

\begin{figure*}[t]
\begin{subfigure}{0.46\linewidth}
	\includegraphics[width = \linewidth]{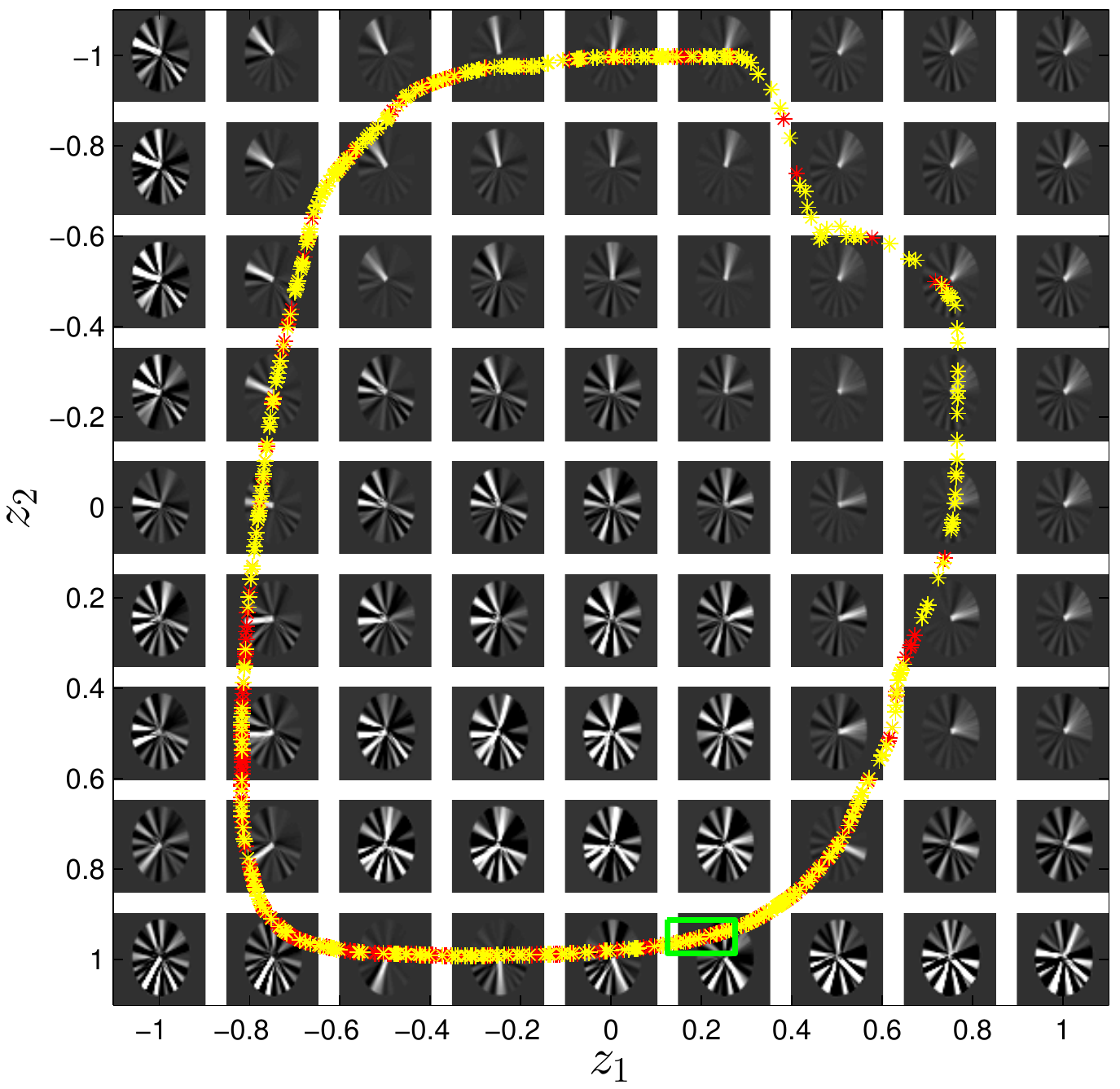}
  \caption{Results from joint learning.}
	\label{fig:latent}	
\end{subfigure}
\hfill
\begin{subfigure}{0.46\linewidth}
	\includegraphics[width = \linewidth]{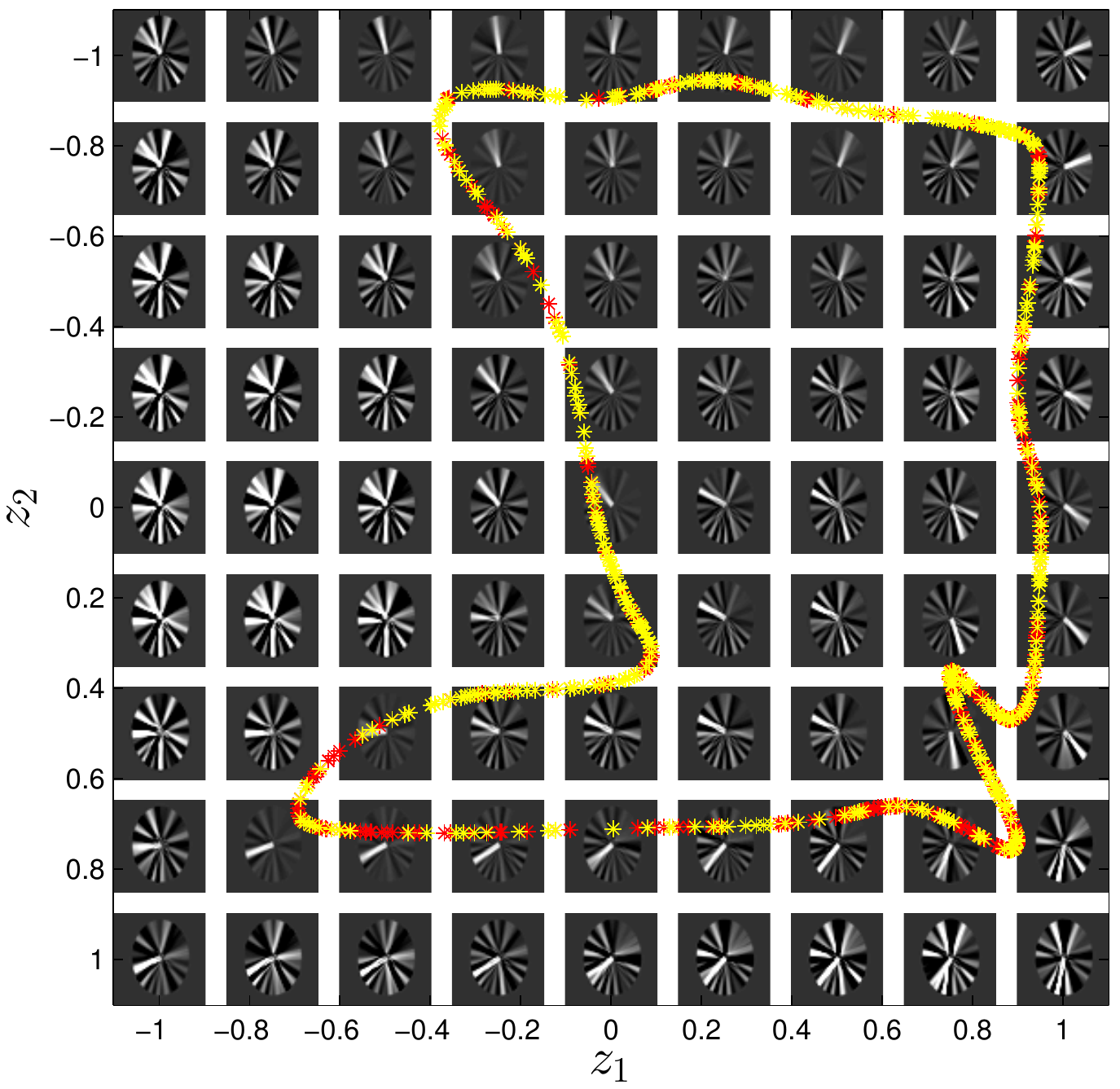}
	\caption{Results from separate learning.}
	\label{fig:latent_sep}
\end{subfigure}
\caption{Exp. 1: The feature space $\feature \in [-1, 1] \times [-1, 1]$ is divided into $9 \times 9$ grid points. For each grid point the decoded high-dimensional image is displayed. The feature values corresponding to the training (red) and validation (yellow) data are displayed. Feature spaces found by joint  (\subref{fig:latent}) and separate (\subref{fig:latent_sep}) parameter learning. A zoomed in version of the green rectangle is presented in \fig\ref{fig:latent_example}.}
	\label{fig:latentBoth}		
\end{figure*}

\fig\ref{fig:latentBoth} displays the ``decoded'' images corresponding to the latent representations using joint and separate training, respectively. In the joint training the feature values line up in a circular shape enabling a low-dimensional dynamical description, whereas separate training finds feature values, which are not even placed sequentially in the low-dimensional representation. Separate training extracts the low-dimensional representations  without context, i.e., the knowledge that these features constitute a time-series. On the other hand, joint training enables the extraction of features  that can also model the dynamical behavior in a compact manner.

\begin{figure}[tb]
	\includegraphics[width = \hsize]{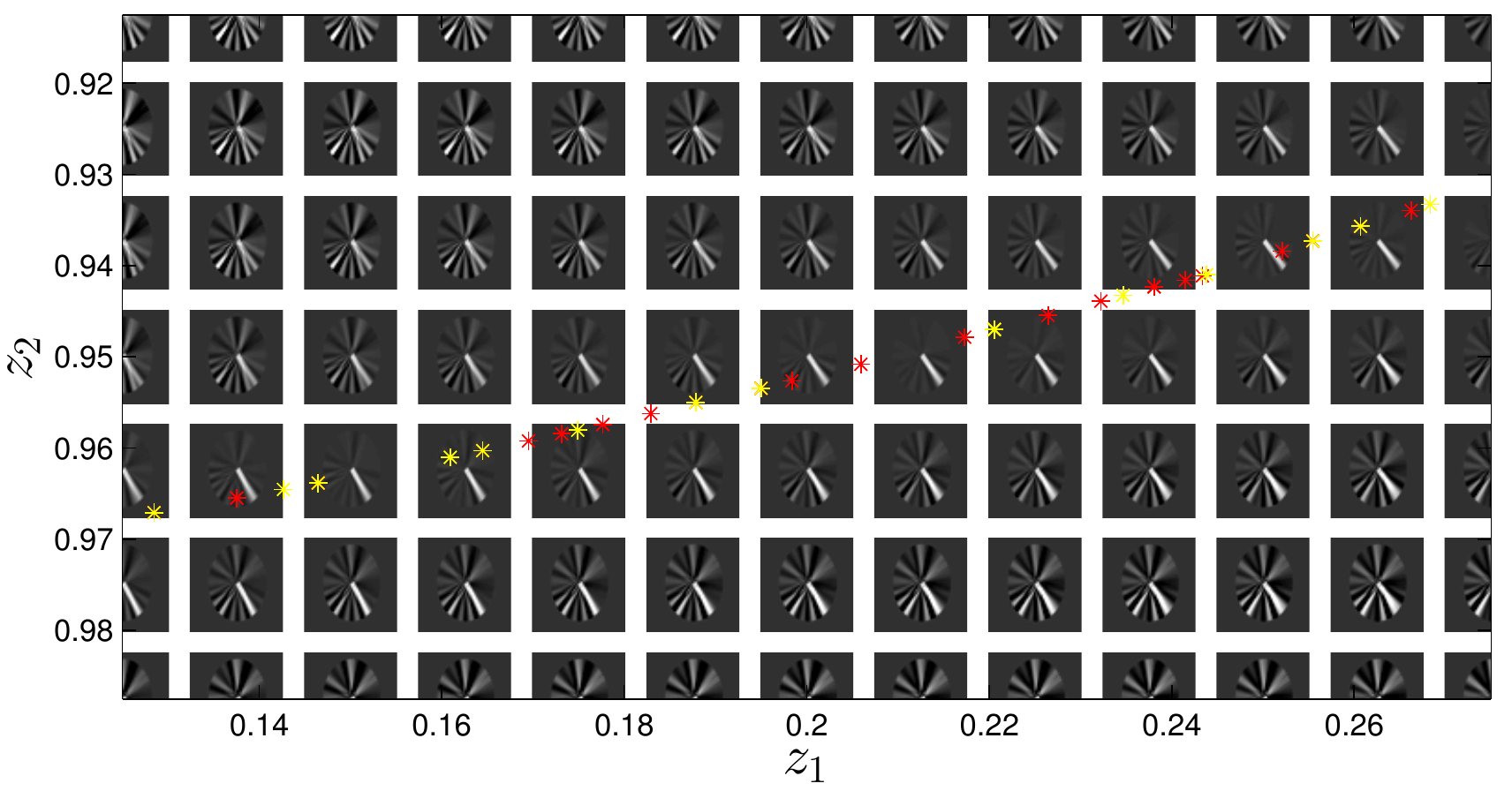}  
\caption{Exp. 1: A zoomed-in version of the feature space, corresponding to the green rectangle in \fig\ref{fig:latent}.}
	\label{fig:latent_example}
\end{figure}

In this particular data set, the data points clearly reside on one-dimensional manifold, encoded by the pendulum angle. However, a one-dimensional feature space would be insufficient since this one-dimensional manifold is cyclic, see \fig\ref{fig:latentBoth}, compare also with the $2\pi$ period of an angle. Therefore, we have used a two-dimensional latent space. Further, only along the manifold in the latent space where the training data reside the decoder produces reasonable outputs. This can be further inspected by zooming in on a smaller region as displayed in \fig\ref{fig:latent_example}.

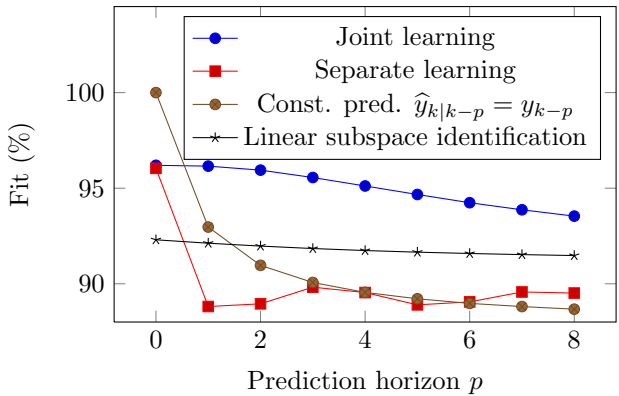
\begin{figure}[tb]
\centering
\begin{tikzpicture}
\pgfsetlinewidth{20}
    \begin{axis}[
        xlabel=Prediction horizon $p$,
        ylabel=Fit ($\%$),
        legend pos= north east,
        ymin=88,ymax=104.5,
        width = \linewidth,
        height = 0.7\linewidth,
    ]    
      \addplot plot coordinates {
(0, 96.1978)
(1, 96.1541)
(2, 95.9451)
(3, 95.5584)
(4, 95.1162)
(5, 94.6709)
(6, 94.2409)
(7, 93.8712)
(8, 93.5397)
    };

   \addplot plot coordinates {
(0, 96.0393)
(1, 88.8115)
(2, 88.9518)
(3, 89.823)
(4, 89.5416)
(5, 88.8983)
(6, 89.0535)
(7, 89.572)
(8, 89.5154)
    };

    \addplot plot coordinates {
(0, 100)
(1, 92.9682)
(2, 90.9634)
(3, 90.0684)
(4, 89.5501)
(5, 89.2161)
(6, 88.9749)
(7, 88.8094)
(8, 88.6715)
    }; 
        \addplot plot coordinates {
(0, 92.301)
(1, 92.123)
(2, 91.9723)
(3, 91.8444)
(4, 91.7392)
(5, 91.654)
(6, 91.5849)
(7, 91.5264)
(8, 91.4775)
    };        
    \legend{Joint learning\\Separate learning\\Const. pred. $\widehat{\meas}_{k|k-p} = \meas_{k-p}$ \\Linear subspace identification\\}
    \end{axis}
\end{tikzpicture}
\caption{Exp. 1: Fitting quality \eqref{eq:fit} for joint and separate learning of features and dynamics for different prediction horizons $p$. The fit is compared with the naive prediction $\widehat{\meas}_{t|t-p} = \meas_{t-p}$, where the most recent image is used for the prediction $p$ steps ahead and a linear subspace-ID method.}
\label{fig:prediction}
\end{figure}
To analyze the predictive performance of the two training methods, we define the fitting quality as
\begin{align} \label{eq:fit}
\text{FIT}_p = 1 - \sqrt{\tfrac{1}{NM}\sum\nolimits_{t=1}^N \|\meas_t - \widehat{\meas}_{t|t-p}\|^2}.
\end{align}
As a reference, the predictive performance is compared with a naive prediction using the previous frame at time step $t-p$ as the prediction at $t$ as $\widehat{\meas}_{t|t-p} = \meas_{t-p}$. The result for a prediction horizon ranging from $p=0$ to $p=8$ is displayed in \fig\ref{fig:prediction}.

Clearly, joint learning (blue) outperforms separate learning in terms of predictive performance for prediction horizons greater than 0. Even by using the last available image frame for prediction (const. pred., brown), we obtain a  better fit than the model that learns its parameter sequentially (red). This is due to the fact that the dynamical model often predicts frames, which do not correspond to any real pendulum, see \fig\ref{fig:results}, leading to a poor fit. Furthermore, joint training gives better predictions than the naive prediction. The predictive performance slightly degrades when the prediction horizon $p$ increases, which is to be expected. Finally we also compare with the subspace identification method~\cite{deMoor:1996} (black, starred), which is restricted to linear models. Such a restriction does not capture the non-linear, embedded features and, hence, the predictive performance is sub-optimal.

\subsection{Experiment 2: Tile moving in the plane}
In this experiment we simulated 601 frames of a moving tile in a $51 \times 51 = 2601$ pixels image. The control inputs are the increments in position in both of the two Cartesian directions. 
As in the previous experiment, the image sequence was reduced to $\text{dim}(\meas_t) = 50$ prior to the parameter learning using PCA. A four layer 50-25-12-8-2 autoencoder was used for feature learning, and an 8-5-2 neural network for the dynamics. 

\begin{figure}[t]
\centering
\begin{subfigure}{\linewidth}
	\includegraphics[width = \linewidth]{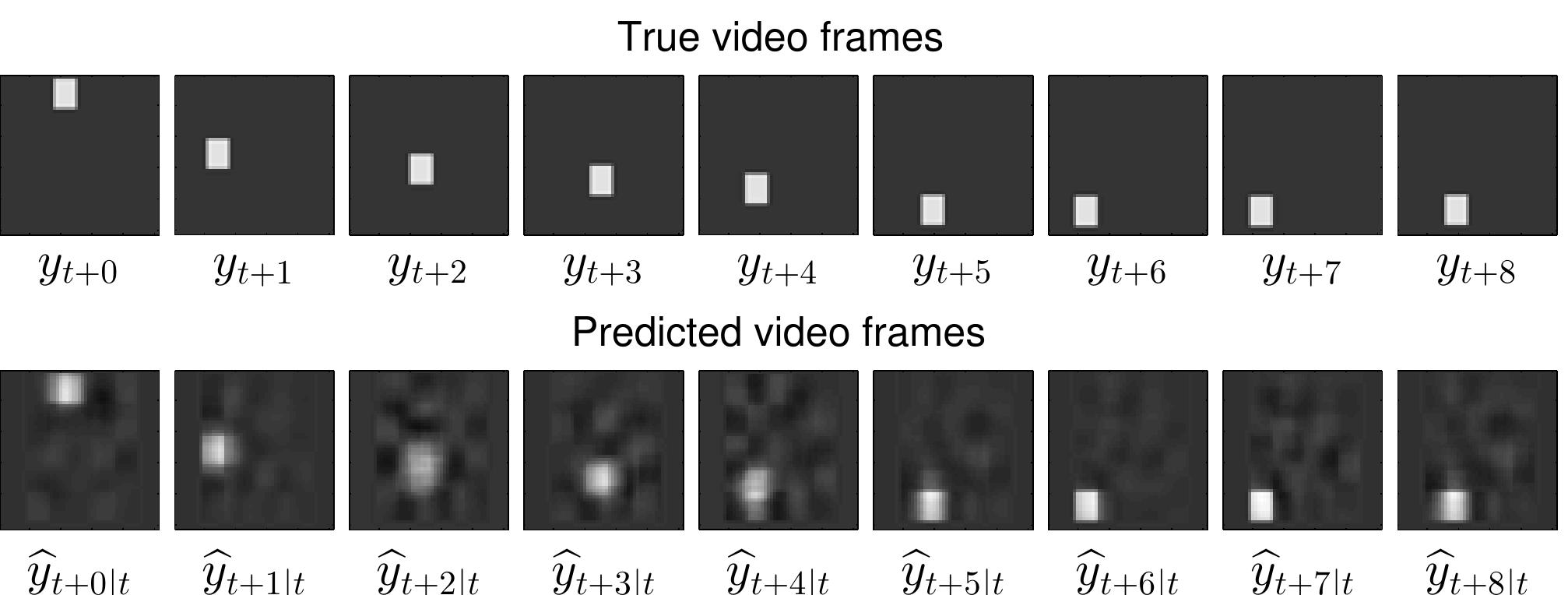}
	\caption{Sequence 1}
    \label{fig:results1_2dtile}
\end{subfigure}\\
\begin{subfigure}{\linewidth}
	\includegraphics[width = \linewidth]{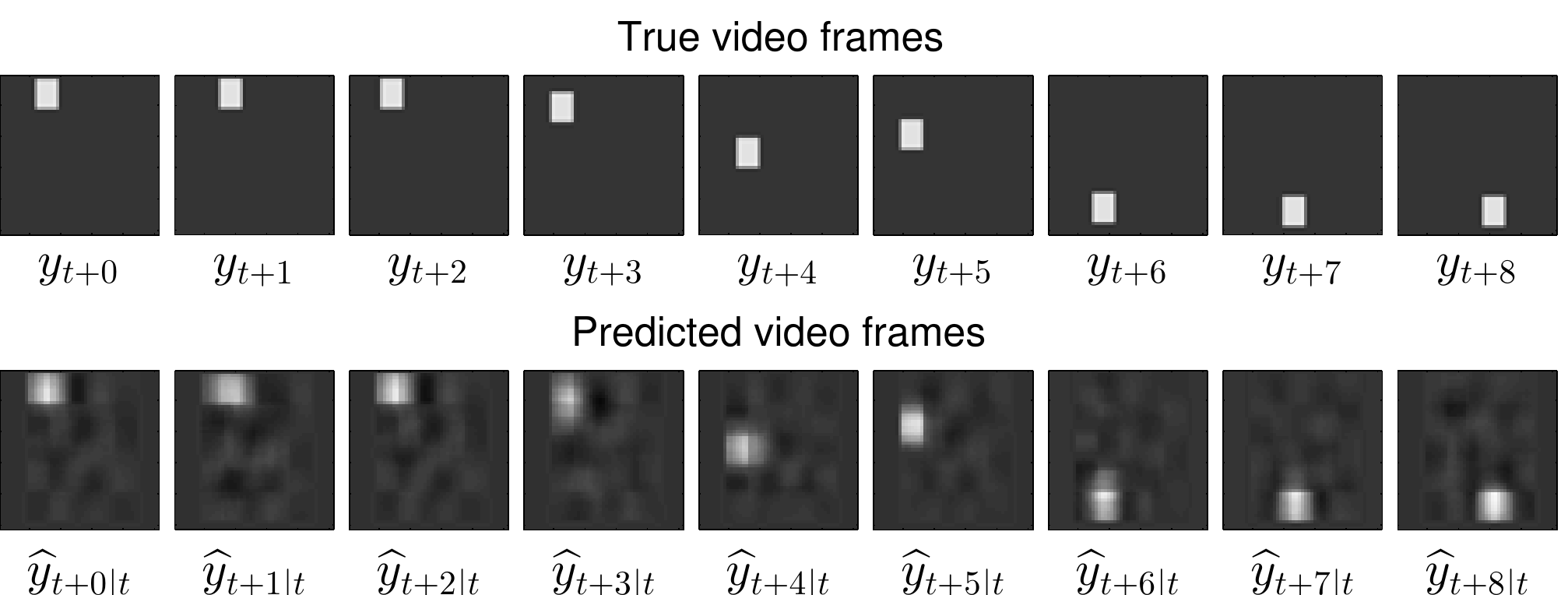}
	\caption{Sequence 2}
	\label{fig:results2_2dtile}
\end{subfigure}
\caption{Exp. 2: True and predicted video frames on validation data.
}
\label{fig:results_2dtile}
\end{figure}

\begin{figure}[t]
	\includegraphics[width = \linewidth]{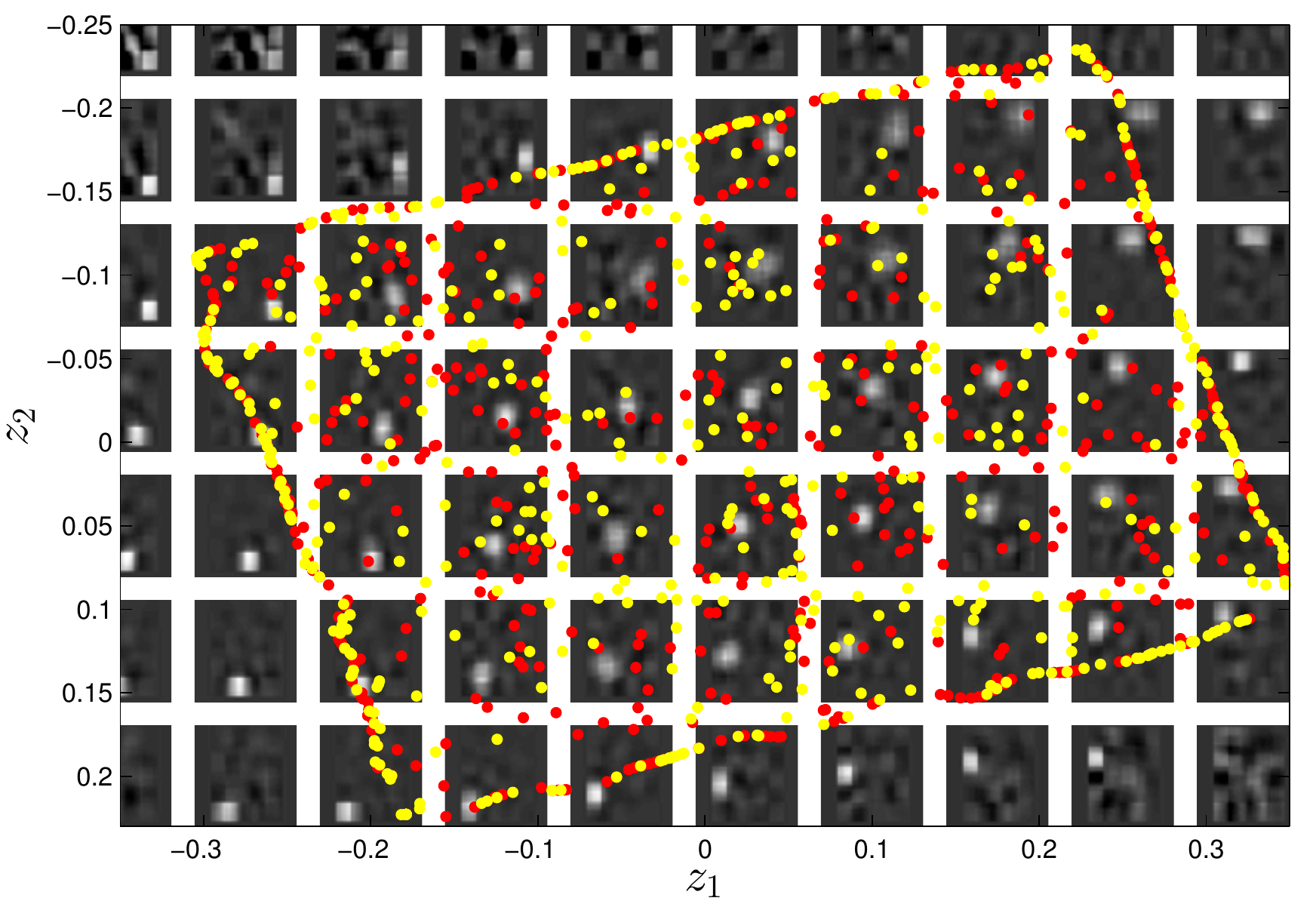}  
\caption{Exp. 2: Feature space zoomed in to the region $\feature \in [-0.35, 0.35] \times [-0.25, 0.25]$. The training (red) and validation (yellow) data reside on a two-dimensional manifold corresponding to the two-dimensional position of the tile.}
	\label{fig:latent_2dtile}
\end{figure}

As in the previous example, we evaluate the performance in terms of long-term predictions. The performance of joint training is illustrated in \fig\ref{fig:results_2dtile} on a validation data set. The model predicts future frames of the tile with high accuracy. In \fig\ref{fig:latent_2dtile}, the feature representation of the data is displayed. The features reside on a two-dimensional manifold encoding the two-dimensional position of the moving tile. The four corners in this manifold represent the four corners of the tile position within the image frame. This structure is induced by the dynamical description. The corresponding feature representation for the case of separate learning does not exhibit such a structure, see the supplementary material.

In \fig\ref{fig:prediction_2dtile}, the prediction performance is displayed, where our model achieves a substantially better fit than naively using the previous frame as prediction.

\section{Discussion}
\label{sec:discussion}

From a system identification point of view, the prediction error method, where we minimize the one-step ahead prediction error, is fairly standard. However, in a future control or reinforcement learning setting, we are primarily interested in good prediction performance on a longer horizon in order to do planning. Thus, we have also investigated whether to additionally include a multi-step ahead prediction error in the cost~\eqref{eq:prediction_error}. These models achieved similar performance, but no significantly better prediction error could be observed either for one-step ahead predictions or for longer prediction horizons.


\begin{figure}[t]
\centering
\begin{tikzpicture}
\pgfsetlinewidth{20}
    \begin{axis}[
        xlabel=Prediction horizon $p$,
        ylabel=Fit ($\%$),
        legend pos= north east,
        ymin=83,ymax=104,
        width = \linewidth,
        height = 0.5\linewidth,
    ]    
      \addplot plot coordinates {
(0, 94.1555)
(1, 94.0422)
(2, 93.868)
(3, 93.6945)
(4, 93.5584)
(5, 93.4499)
(6, 93.3689)
(7, 93.298)
(8, 93.2361)
    };

    \addplot plot coordinates {
(0, 100)
(1, 87.1909)
(2, 86.3216)
(3, 86.0445)
(4, 85.8246)
(5, 85.6132)
(6, 85.5812)
(7, 85.4631)
(8, 85.3723)
    };    
    \legend{Joint learning\\Const. pred. $\widehat{\meas}_{k|k-p} = \meas_{k-p}$\\}
    \end{axis}
\end{tikzpicture}
\caption{Experiment 2: The fit \eqref{eq:fit} for different prediction horizons $p$.}
\label{fig:prediction_2dtile}
\end{figure}
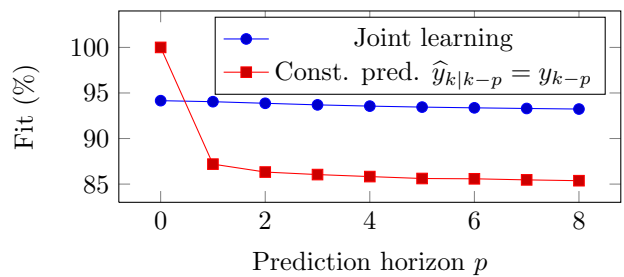

Instead of computing the prediction errors in image space, see \eqref{eq:prediction_error}, we can compute errors in feature space  to avoid a decoding step back to the high-dimensional space according to~\eqref{eq:prediction_mapping}. However, this will require an extra penalty term in order to avoid trivial solutions that map everything to zero, eventually resulting  in a more complicated  and less intuitive cost function.

Although joint learning aims at finding a feature representation that is suitable for modeling the low-dimensional dynamical behavior, the pre-training initialization as described in Section~\ref{sec:pretraining} does not. If this pre-training yields feature values far from ``useful'' ones for modeling the dynamics, joint training might not find a good model.

The network structure has to be chosen before the actual training starts. Especially the dimension of the latent state and the order of the dynamics have to be chosen by the user, which requires a some prior knowledge about the system to be identified. In our examples, we chose the latent dimensionality based on insights about the true dynamics of the problem. In general, a model selection procedure will be preferable to find both a good network structure and a good latent dimensionality.

\section{Conclusions and future work}
\label{sec:conclusions}

We have presented an approach to non-linear system identification from high-dimensional time series data. Our model combines techniques from both the system identification and the machine learning community. In particular, we used a deep auto-encoder for finding low-dimensional features from high-dimensional data, and a nonlinear autoregressive exogenous model was used to describe the low-dimensional dynamics. 

The framework has been applied to a pendulum moving in the horizontal plane. The proposed model exhibits good predictive performance and a major improvement has been identified by training the auto-encoder and the dynamical model jointly instead of training them separately\slash sequentially.

Possible directions for future work include (a) robustify learning by using denoising autoencoders \cite{Vincent2008} to deal with noisy real-world data (b) apply convolutional neural networks, which are often more suitable for images; (c) exploiting the learned model for learning controller purely based on pixel information; (c) Sequential Monte Carlo methods will be investigated for systematic treatments of such nonlinear probabilistic models, which are required in a reinforcement learning setting.

In a setting where we make decisions based on (one-step or multiple-step ahead) predictions, such as optimal control or model-based reinforcement learning,  a probabilistic model is often needed for robust decision making as we need to account for models errors~\cite{Schneider1997,Deisenroth2014}.  An extension of our present model to a probabilistic setting is non-trivial since random variables have to be transformed through the neural networks, and their exact probability density functions will be intractable to compute. Sampling-based approaches or deterministic approximate inference are two options that we will investigate in future.

\subsection*{Acknowledgments}
This work was supported by the Swedish Foundation for Strategic Research under the project \emph{Cooperative Localization} and by the Swedish Research Council under the project \emph{Probabilistic modeling of dynamical systems} (Contract number: 621-2013-5524). MPD was supported by an Imperial College Junior Research Fellowship.

\appendix

\section*{Supplementary material}
\begin{figure}[h]
	\includegraphics[width = 0.8\linewidth]{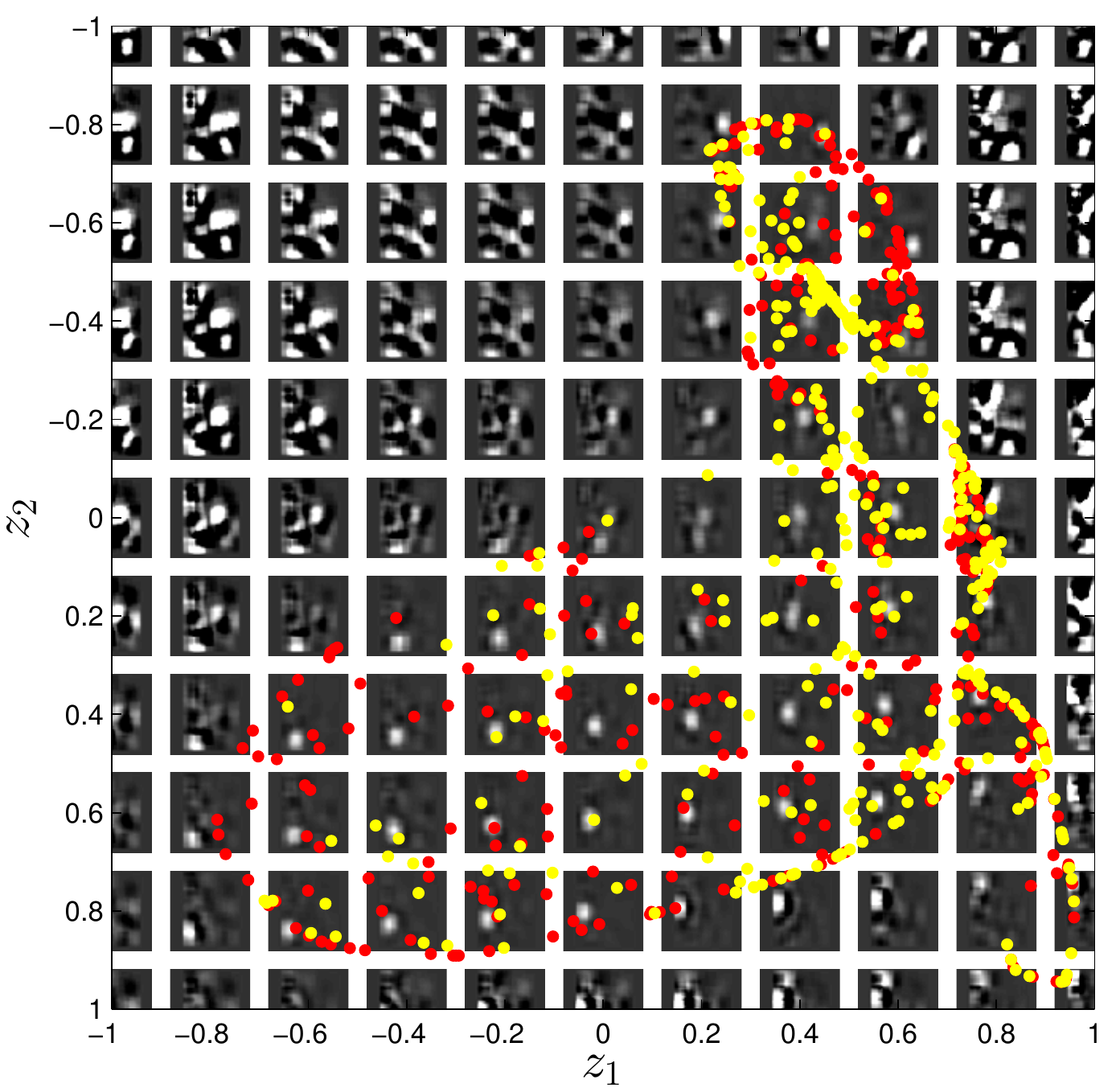}  
\caption{Exp. 2: The feature space in the region $\feature \in [-1, 1] \times [-1, 1]$ for separate learning. The training (red) and validation (yellow) data does not reside on a structured a two-dimensional tile-formed manifold as it was the case for the joint learning in Fig.~\ref{fig:latent_2dtile}.}
	\label{fig:latent_2dtile_sep}
\end{figure}

In the second experiment in the paper, results the joint learning of prediction and reconstruction error have been reported. The joint learning brings structure to feature values, which is not present if the autoencoder is learn separately, see Fig.~\ref{fig:latent_2dtile_sep}. 



\end{document}